\documentclass[runningheads]{llncs}

\usepackage{eccv}

\usepackage{eccvabbrv}

\usepackage{graphicx}
\usepackage{booktabs}
\usepackage{arydshln}
\usepackage{xfakebold}

\usepackage{colortbl}
\definecolor{lightgray}{gray}{0.94}

\usepackage[accsupp]{axessibility}  % Improves PDF readability for those with disabilities.

\usepackage{hyperref}

\usepackage{orcidlink}

\begin{document}

% \title{Progressive Masking Guided Diffusion for Unsupervised Image Translation and Anomaly Detection} 
\title{Source-Agnostic Image Translation Based on Latent Aware Adaptive Masking} 
\titlerunning{Latent Aware Adaptive Masking}

\author{Tomislav Dobrički\inst{1}\orcidlink{0009-0007-8427-3469} \and
Byung-Woo Hong\inst{1}\orcidlink{0000-0003-2752-3939} }

\authorrunning{T. Dobrički and B.-W. Hong}

\institute{Chung-Ang University, Seoul, South Korea \\
\email{\{tomislav, hong\}@ai.cau.ac.kr} %\\
% Project webpage: \url{https://github.com/dtoma95/PM-Edit}
}
\maketitle

\begin{abstract}
In this work, we propose a source-agnostic framework that dynamically refines a binary mask throughout the reverse diffusion process by computing the discrepancies of a pretrained diffusion model's prediction for each latent time step. Rather than relying on a fixed threshold, our method introduces a time-dependent statistical thresholding scheme derived from the empirical mean and standard deviation of prediction discrepancies across the latent noisy images from the target distribution. This allows the mask to adapt to the model's varying predictive confidence at different noise levels, effectively isolating domain-specific regions while preserving global structural coherence. Experimental results on the AFHQ and Celeba-HQ datasets demonstrate that our approach outperforms state-of-the-art unsupervised Image-to-Image methods in both realism (FID, KID) and faithfulness (SSIM, LPIPS). By requiring only a pretrained model of the target domain, our approach enables precise, automated localization and seamless translation across diverse source distributions without any specialized training. The project source code is available at: \url{https://github.com/dtoma95/PM-Edit}
\end{abstract}

% \let\thefootnote\relax\footnotetext{The project source code is available at: \url{https://github.com/dtoma95/PM-Edit}.}
% Use it in your document text:
% \blfootnote{This research was funded by the National Science Foundation.}
% \blfootnote{The project source code is available at: \url{https://github.com/dtoma95/PM-Edit}.}
\section{Introduction}
\label{sec:intro}

Image-to-Image (I2I) translation is a fundamental task in computer vision that has many possible applications in fields such as medical imaging and industrial anomaly detection. However, traditional I2I translation methods rely on predefined pairs of source and target data samples for training~\cite{isola2017image, zhu2017toward, wang2018high}, making them highly impractical for real world applications where image pairs are often unavailable. For this reason, unsupervised I2I translation methods~\cite{huang2018multimodal,liu2017unsupervised,radford2015unsupervised} present great potential, as they do not require corresponding pairs from the source and target domains, alleviating the need for extensive data collection.

Previous approaches leverage the power of Generative Adversarial Networks (GANs)~\cite{goodfellow2014generative, karras2020analyzing}. Methods such as CycleGAN~\cite{zhu2017unpaired} use GANs to learn a bidirectional mapping between a source and target domain. Later, CUT~\cite{park2020contrastive} leverages contrastive learning to maximize the mutual information between the input and output images. Allowing the neural network to isolate domain-specific features of the target dataset and ignore the shared features. 
Later methods improve on these works~\cite{wang2023unsupervised,torbunov2023uvcgan,li2018unsupervised, han2021dual}, but are limited by the capabilities of GANs. 

\begin{figure}[tb]
  \centering
  \includegraphics[width=\linewidth]{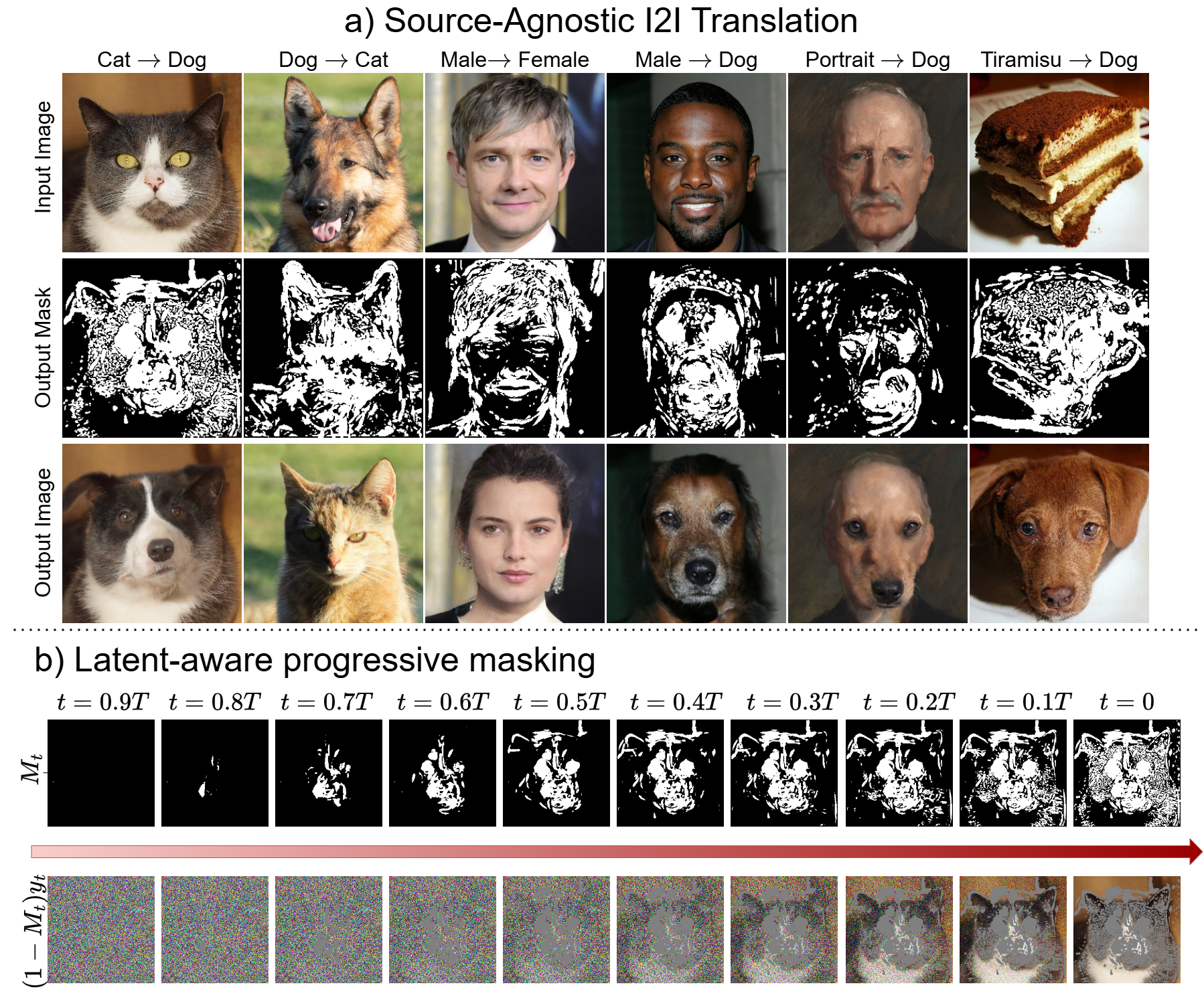}
  \caption{a) For a given source image (top), masks are computed at each time step (middle; only final mask is shown) and the translated image (bottom) is generated by inpainting. The generative network is only trained on the target domain in each example. Note that the displayed masks do not directly correspond to the image modifications because the masks are slightly dilated and blurred during inpainting. b) The mask at each time step (top) adapts to the latent noise level. When applied to the noised image (bottom) the mask conceals features foreign to the target domain.
  % We present a progressive masking based approach for source domain agnostic image editing. We display the mask applied to the noised source image (bottom). More prominent features need to be masked earlier into the backwards diffusion process, while finer details emerge only at lower noise levels. At each time steps the mask makes the image indistinguishable for the target domain, which is "dog" in this example.
  }
  \label{fig:intro}
\end{figure}

Denoising Diffusion Probabilistic models (DDPMs~\cite{song2020denoising, rombach2022high, dhariwal2021diffusion, nichol2021improved, ho2020denoising} are a class of generative methods that achieve state-of-the-art results in image generation. They achieve this via an iterative process where images are perturbed using Gaussian noise in a forward process, and then a neural network is trained to perform a reverse denoising process to generate new images that belong to the desired distribution. Because of their superior generative capabilities, there has been a lot of interest in utilizing DDPM for unsupervised I2I translation~\cite{sasaki2021unit}. Recent methods such as EGSDE~\cite{zhao2022egsde} and CycleDiffusion~\cite{wu2023latent} aim to guide the denoising process of pretrained DDPM networks by injecting knowledge of the differences between the features of the source and target domains. Older diffusion guidance methods such as SDEdit~\cite{meng2021sdedit} and ILVR~\cite{choi2021ilvr} are source-agnostic, meaning that they can be applied to images of any domain given a generator from the target. However, they often fail in either faithfulness or realism, as they are not adaptive to different input images. 

Mask guided diffusion has been used by several methods in other domains, such as image editing~\cite{couairon2022diffedit} and anomaly detection~\cite{yao2024glad}. These approaches automatically designate a masked region by comparing the discrepancy in the neural networks prediction in the latent space. This discrepancy is then binarized using a fixed threshold. An output image is then generated by diffusion inpainting, where the unmasked regions guide the denoising process to create a cohesive final image. However, the masks in these methods are computed at a single latent time step and remain constant afterward. Additionally, the neural network is typically pretrained using knowledge about the source distribution.

We propose a novel, source-agnostic masking approach for unsupervised I2I translation. In our method, a mask is computed using the latent variable of a pretrained DDPM neural network at each noise level; meaning that it progressively grows larger as more details of the image are revealed. Our approach assumes that for every time step, there exists a corresponding minimal mask required to occlude domain-specific regions from the noise estimator. Translation is then achieved via inpainting, where the appropriate mask is applied at its corresponding time step in the DDPM backward process. This means that features requiring large modifications are altered early, while finer details can be altered at the end of the generation. Our approach relies only on a denoising neural network that is trained to generate images of the target domain, and the training process is not different from the typical DDPM methods.

Similar to the previously mentioned mask guidance-based methods, we compute the mask by comparing the difference in the denoising network's prediction between the noised source image and our generated sample. This is achieved using a discrepancy metric applied to the model's estimation of the denoised image. Unlike previous works, we compute the binary mask using a threshold that corresponds to each noise level. This is achieved by pre-computing the discrepancy statistics of the pretrained network at each time step, using samples from the target domain. To evaluate the effectiveness of this approach, we evaluate our method on three standard image translation tasks using the CelebaHQ~\cite{karras2017progressive}, and AFHQ~\cite{choi2020stargan} datasets. Our method outperforms the state-of-the-art in unsupervised I2I translation on several metrics of both faithfulness and realism. A demonstration of the source-agnostic I2I translation capabilities, together the latent noise aware masking process is shown in Figure~\ref{fig:intro}.
In summary, the primary contributions of this work are as follows:
\begin{itemize}
\item We introduce Progressive Mask Editing (PM-Edit): an unsupervised I2I translation approach based on an adaptive masking framework that guides the diffusion model by applying a progressively growing minimal mask at each latent time step. This approach is source agnostic, in that the denoising neural network is trained solely on the target distribution. The masks are computed using the prediction discrepancy of the network.
% between the noised source image and the generated sample.
\item A time-dependent statistical thresholding scheme. In order to compute the masks, we first calculate the statistical properties of the neural networks prediction discrepancies on samples from the target distribution. This allows for robust masking at each latent noise level and can be tuned using a single hyperparameter.
\item Experiments that demonstrate superior results over state-of-the-art methods both qualitatively and quantitatively, evaluated on standard I2I translation tasks on the AFHQ, CelebaHQ and PIE-Bench datasets. In addition, we provide ablation studies for different hyperparameter settings of our method.
\end{itemize}

\section{Related Work}

\subsection{Diffusion Models}
Diffusion probabilistic models are a type of generative model that can generate realistic images of a given domain by reversing an iterative noise degradation process \cite{sohl2015deep}. Denoising Diffusion Probabilistic models (DDPM) \cite{ho2020denoising} simplified the formulation by training a denoising neural network to reverse the diffusion process. Later works improved on this formulation and achieved impressive results on a variety of image synthesis tasks\cite{nichol2021improved, dhariwal2021diffusion, rombach2022high}. Denoising diffusion implicit models (DDIM) \cite{song2020denoising} improve the sampling efficiency of diffusion models by employing a non-Markovian degradation process. This also introduces a deterministic alternative for DDPM, as previous approaches relied on a stochastic backwards process where random noise is injected at each sampling step. This deterministic approach allows for the inverse mapping of a given image from the target domain to its corresponding noise latent. Other than image generation, diffusion models have been adapted to several other computer vision problems, such as image editing~\cite{couairon2022diffedit, kawar2023imagic, gandikota2024unified}, anomaly detection~\cite{zhang2025diffusionad, yao2024glad, wolleb2022diffusion} or image inpainting~\cite{lugmayr2022repaint, zhang2023towards}.
 
\subsection{Unsupervised Image to Image Translation}
Early works have shown reasonable results in unsupervised (or unpaired) I2I translation by utilizing GANs. Works such as CycleGAN~\cite{zhu2017unpaired}, DualGAN~\cite{yi2017dualgan} and DiscoGAN~\cite{kim2017learning}define the translation as a bidirectional mapping between the source and target domains via cycle-consistency, and subsequent methods improve on this approach~\cite{li2018unsupervised, kim2017learning, kim2019u}. To avoid the limitations of bijection transformation, subsequent works aim for a one-sided mapping approach. GCGAN~\cite{fu2019geometry} enforces geometric consistency while DistanceGAN~\cite{benaim2017one} learn a mapping to maintain a distance between the input and output images. Finally, CUT~\cite{park2020contrastive} achieves stat-of-the-art the results for GAN-based methods by using contrastive learning to ensure patch-wise mutual information.

Given the superior generative capabilities of Diffusion-based frameworks, many unpaired I2I translation approaches aim to take advantage of them to create realistic translations. UNIT-DDPM~\cite{sasaki2021unit} and CycleDiffusion~\cite{wu2023latent} utilize two denoising neural networks using cycle-consistency. EGSDE~\cite{zhao2022egsde} guides the diffusion model using an energy function pretrained on the source and target domains. At the same time, Brownian Bridge Diffusion Models (BBDM)~\cite{li2023bbdm}  and Unpaired Schrödinger Bridge (UNSB)~\cite{kim2023unpaired} deviate from standard diffusion by mapping the forward and backward process between the source and target domain directly, instead of starting the inference process and Gaussian noise. In terms of source agnostic approaches ILVR~\cite{choi2021ilvr} and SDEdit~\cite{meng2021sdedit} use reference models to guide the DDPM generative process. However these methods require careful tuning of hyperparameter and modify image regions indiscriminately, without the consideration of domain-specific and unspecific features. Our approach is different as we utilize an adaptive masking scheme to achieve source-agnostic I2I translation. And unlike previous masking-based approaches like DiffEdit~\cite{couairon2022diffedit}, we do not use a constant threshold or mask.

 \section{Preliminaries}
\subsection{Denoising Diffusion Models}
Denoising diffusion probabilistic models (DDPM)~\cite{sohl2015deep, ho2020denoising, nichol2021improved} are a family of iterative generative models that consist of a forward and a backward process. In the forward process, data samples are gradually perturbed over $T$ time steps until they are indistinguishable from Gaussian noise. For a data sample $x_0$, its noised version at any arbitrary time step $t$ is given by:
\begin{equation}
\label{forward_process}
x_t = \sqrt{\alpha_t}x_0 + \sqrt{1-\alpha_t}\epsilon,
\end{equation}
where the coefficient $\alpha_t$ corresponds to the noise level as defined by a noising schedule, and $\epsilon$ is randomly sampled as $\epsilon \sim \mathcal{N}(0,I)$. The noising schedule is a decreasing function with respect to $t$, such that $\alpha_0 = 1$ and $\alpha_T \approx 0$.

The goal of the backward process is to reverse the perturbation by iteratively denoising the sample, starting from the final time step. This is modeled using a deep neural network trained to predict the added noise according to the current time step. During training, the neural network parameters $\theta$ are optimized to minimize the objective:
\begin{equation}
\label{ddpm_loss}
L(\theta) = \mathbb{E}_{x_0, \epsilon, t} \left[ \left\| \epsilon - \epsilon_\theta\left(x_t, t\right) \right\|_2^2 \right].
\end{equation}
Inference is performed by first sampling $x_T \sim \mathcal{N}(0,I)$ and then iteratively denoising the image for $T$ time steps until the generated sample $x_0$ is provided. This generative process is notoriously slow, as diffusion models require a large value for $T$ to generate realistic images.

Denoising Diffusion Implicit Models (DDIM)~\cite{song2020denoising} improve the sampling of DDPM by computing a prediction of $x_0$ at every $t$. For convenience, we denote this projection as the function $f_\theta$:
\begin{equation}
\label{ddim_project}
f_\theta(x_t, t) := \frac{x_t - \sqrt{1 - \alpha_t} \epsilon_\theta(x_t, t)}{\sqrt{\alpha_t}}.
\end{equation}
Noise is then re-injected to correspond to the next step of the backward process:

\begin{equation}
\label{ddim_step}
x_{t-1} = \sqrt{\alpha_{t-1}}f_\theta(x_t, t)  + \sqrt{1 - \alpha_{t-1} - \sigma_t^2} \epsilon_\theta(x_t, t) + \sigma_t \epsilon_t,
\end{equation}
where $\sigma_t$ is the standard deviation of the stochastic noise added at each step, and if it is set to 0, the generative process becomes deterministic. DDIM introduces a more flexible inference process, allowing for faster sampling because eq.~\ref{ddim_step} does not have to be tied to the pre-defined number of time steps.
\subsection{Mask Guided Diffusion}
\label{sec:mask_guided_diffusion}
% Although, mask guided diffusion is utilized in both image editing and anomaly detection applications, we argue that the current approaches fall short in both cases. 
 We consider a pre-trained generative denoising neural network $\epsilon_\theta$ designed to generate images of a given target domain  $\mathcal{X} \subset \mathbb{R}^{C\times H \times W}$.  For a reference image $y$ from a potentially unknown source distribution $\mathcal{Y} \subset \mathbb{R}^{C\times H \times W}$, we assume that there is a mask $M \in \{0, 1\}^{H \times W}$ that identifies regions of $y$ that are inconsistent with $\mathcal{X}$. Regions where $M=1$ contain evidence that the image does not belong to the target domain; $M=0$ signifies the domain-consistent regions. 
 % In the context of anomaly detection, $M$ represents the mask segmentation of the anomalous areas of the anomalous image $y$ with respect to the domain of normal images $\mathcal{X}$. 

In SDEdit~\cite{meng2021sdedit}, the generated image $x_0$ is conditioned to resemble $y$ by initializing the inference process at an intermediate time step $t_{start}$. Subsequently, DiffEdit~\cite{couairon2022diffedit} utilizes a mask to guide the diffusion process and minimize unwanted modifications to the reference image. This is equivalent to diffusion inpainting, where the update step is expressed as:

\begin{equation}
\label{masked_diff}
\tilde{x}_{t} = Mx_{t} + (1-M)y_t,
\end{equation}
where $y_t$ is obtained by applying the forward diffusion from eq.~\ref{forward_process} on the reference image, and $x_t$ is sampled from the denoising network inference process. In DiffEdit, $M$ is computed at
the predetermined starting time step $t_{start}$ by comparing the discrepancy between the noise predictions of $\epsilon_\theta$ for different class condition inputs and binarized according to a fixed threshold.

% Prior works in anomaly detection leverage the generative prior of diffusion models for anomaly detection by adding a certain amount of noise to $y$ and reconstructing its non-anomalous version by running the diffusion generative process. The anomaly map is then determined using a similarity measure between the original image and the final reconstruction. Methods like GLAD~\cite{yao2024glad} compute a mask by comparing the discrepancies between $f_\theta(x_t, t)$ and $f_\theta(y_t, t)$ to better determine $t_{start}$ and guide the generative process.  However, computing $M$ at a single point may mean that smaller anomalies are overlooked if several anomalous regions are present in $y$. 

\section{Methodology}
In this section we first introduce our latent aware adaptive masking framework, then we specify the precomputation process for the dynamic threshold. And at the end we detail the inpainting methodology used to generate realistic images. An overview of our method is given in Figure~\ref{fig:prog_mask_graph}.
\subsection{Latent Aware Adaptive Masking}
\label{sec:progmask}
We propose an adaptive masking framework to address the limitations inherent in using a static, single-step computation for $M$. Our approach is motivated by the observation that the optimal mask is not constant across the diffusion process; rather, for every noise level $t$, there exists a corresponding minimal mask $M_t$ required to effectively occlude domain-specific regions from the noise estimator. By iteratively refining this mask, we ensure that the guidance remains sensitive to the evolving structural details of the image as it emerges from noise.

% We propose a progressive mask guidance framework to overcome the downsides of a single computation of $M$. We argue that for every noise level $t$, there is a corresponding minimal mask $M_t$ such that it can hide the domain-specific regions from the noise estimator network. 
To determine which regions to mask at a given time step $t$, we rely on the difference in the predictions of the pretrained denoising neural network. Projections of the reference and reconstructed images are compared using a pixel wise difference metric. For simplicity, we define this difference $D_t$ to be the element-wise L1 distance:

\begin{figure}[tb]
  \centering
  \includegraphics[width=\linewidth]{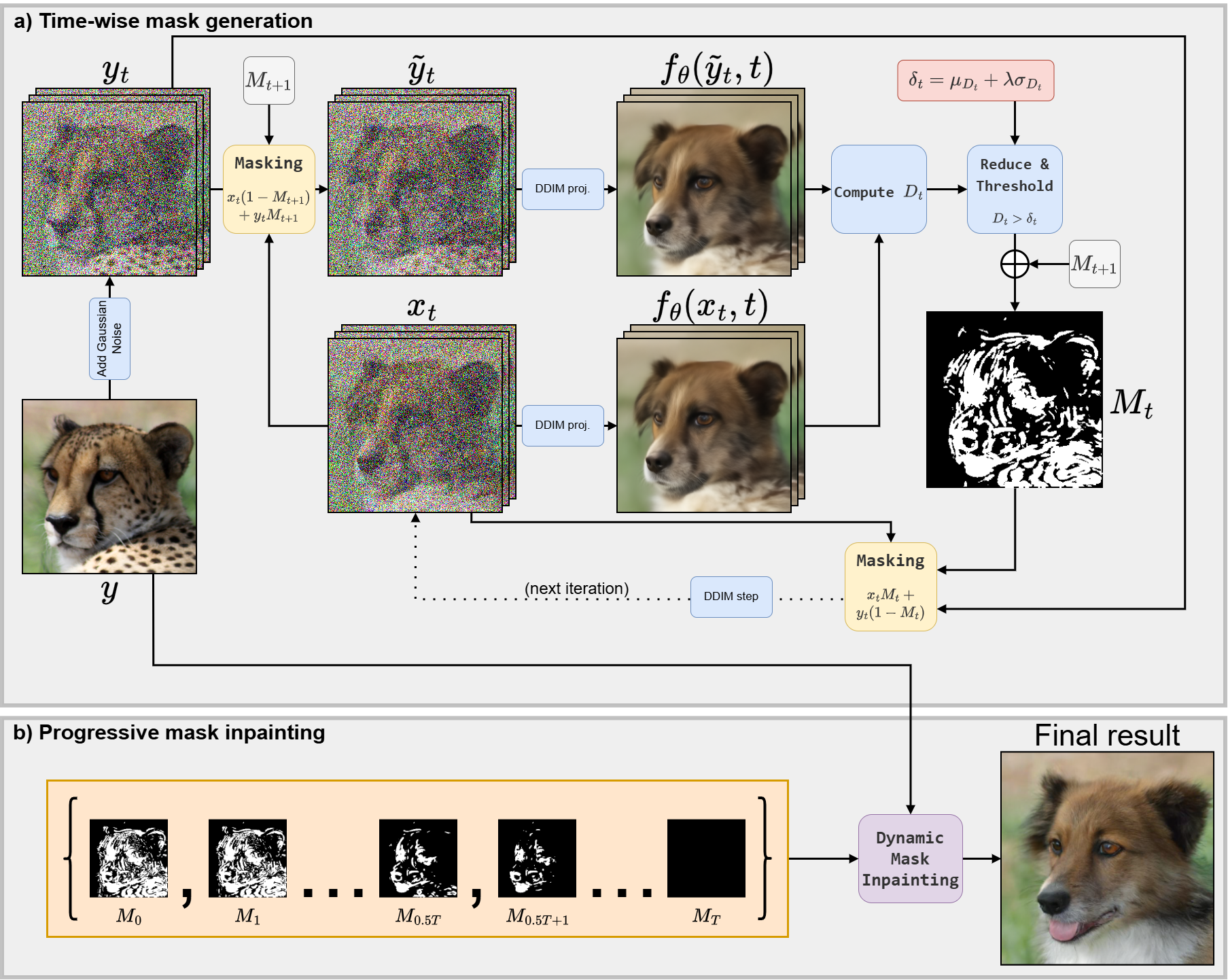}
  \caption{
  Overview of our image editing method. a) For a given reference image $y$ an appropriate mask is computed at each time-step of the DDIM forward process. The current mask $M_t$ is computed by the union between the previous mask and the discrepancy between the model's prediction for the noisy masked reference image $\tilde{y}_t$ and the generated sample $x_t$. b) The output list of masks is used to perform inpainting to generate a more realistic final result. Any diffusion-based inpainting algorithm can be used. At each time step $t$ the appropriate mask $M_t$ is applied to guide generation.}
  
  \label{fig:prog_mask_graph}
\end{figure}

\begin{equation}
\begin{split}
\label{mask_comp}
D_{t}(x_t, y_t) &:= \left\lvert f_\theta(x_t, t) - f_\theta(y_t, t)\right\rvert.
% M_{t} &= ||f_\theta(x_t, t) - f_\theta(\tilde{y}_t, t)||_2^2 < \delta_t,
\end{split}
\end{equation}
The latent variable $x_t$ is generated using the deterministic DDIM sampling applied to its masked version in the previous time step:
\begin{equation}
\begin{split}
\label{prog_mask_inf}
\tilde{x}_{t+1} &= M_{t+1}x_{t+1} + (1-M_{t+1})y_{t+1}, \\
x_t &= \sqrt{\alpha_{t}}f_\theta(\tilde{x}_{t+1}, t+1)  + \sqrt{1 - \alpha_{t}} \epsilon_\theta(\tilde{x}_{t+1}, t+1).
\end{split}
\end{equation}
% Because in eq~\ref{mask_comp}, the $\tilde{y}_t$ and $x_t$ differ only in the masked regions of the image; this allows the model to ignore the discrepancies in regions that were already detected as anomalous during previous steps. For this reason, we derive the mask $M_t$ as the union of the previous mask and the binarized distance $D_t$:
Finally, we define the mask $M_t$ as the union of the previously accumulated mask $M_{t+1}$ (with $M_T$ initialized as 0 for all pixels) and the mask that demarcates the pixels that need to be obscured in the current step $\Delta M_t$: 
\begin{equation}
\begin{split}
\label{prog_mask_accumulation}
\tilde{y}_{t} &= M_{t+1}y_t + (1-M_{t+1})x_t, \\
\Delta M_t &= \left(D_t\left(x_t, \tilde{y}_t\right) > \delta_t\right),
\end{split}
\end{equation}
where $\delta_t$ is a scalar time-dependent threshold value. Finally, we define the mask $M_t$ as the union of the previous mask $M_{t+1}$ and computed discrepancy:
\begin{equation}
M_t = (M_{t+1} + \Delta M_t \geq 1).
\end{equation}
% \begin{equation}
% % M_t \\
% M_t =\max(M_{t+1},D_t > \delta_t).
%   % \begin{cases}
%   %   1 & max(M_{t+1},D_t) > \delta_t\\
%   %   0 & D_t \leq \delta_t
%   % \end{cases}
% \end{equation}
% M_{t} &= ||f_\theta(x_t, t) - f_\theta(\tilde{y}_t, t)||_2^2 < \delta_t

Because $x_t$ is derived from the masked state $\tilde{x}_{t+1}$, the two images differ only in regions bounded by $M_{t+1}$. By comparing their respective projections via $f_\theta$, we isolate the model's prediction error caused specifically at the transition from $t+1$ to $t$. This allows the framework to ignore discrepancies in regions previously identified as anomalous and focus on newly emerging deviations. To ensure that the discrepancies are limited to the denoising network's prediction error, we sample the initial fake image $x_T$ and the noised reference images at every time step $y_t$ using the same noise $\epsilon_{const} \sim \mathcal{N}(0,I)$. However, even with a deterministic sampling process, the diffusion model predictions can still have large outlier values, especially at larger time steps where noise dominates and the network is less certain in its predictions. To counteract this, we determine $D_t$ as an average over $N$ batches of images with different initial noise.
\subsection{Time-Dependent Threshold}

Determining an adequate value for the threshold $\delta_t$ is not trivial, as it depends on many factors, such as the noise schedule and training procedure of the neural network $\epsilon_\theta$. Previous works often use an arbitrary, predetermined value for the threshold. This is not ideal, as it is expected that the statistical properties of $D_t$ largely vary between time-steps. This makes the choice of a single constant threshold highly undesirable, as it limits the flexibility of the masking approach. 

% For this reason, we chose to establish the optimal values of $\delta_t$ empirically. We aim to determine the expected prediction deviation at every time step. First, we generate a set of images from the target dataset using the typical DDIM generative process. Then we repeatedly compute the distances using the chosen dissimilarity metric and calculate the mean $\mu_{D_t}$ and standard deviation $\sigma_{D_t}$ over a total of 10k samples for each time-step. For a given image from the target dataset $z$:
% \begin{equation}
% D_{t} = ||f_\theta\left(\sqrt{\alpha_{t}}f_\theta\left(z_{t+1}, t+1\right)  + \sqrt{1 - \alpha_{t}} \epsilon_\theta\left(z_{t+1}, t+1\right), t\right) - f_\theta\left(z_t, t\right)||_2^2,
% \end{equation}
% where $z_t$ and $z_{t+1}$ are given via Eq.~\ref{forward_process}.
% The threshold is then determined using the formula:
% \begin{equation}
% \delta_t = \mu_{D_t} + \lambda\sigma_{D_t},
% \end{equation}

For this reason, we establish the optimal values of $\delta_t$ empirically by computing the expected prediction deviation at every time step. We sample images from the target dataset $z \sim \mathcal{X}$ in order to compute the expected discrepancy of the networks prediction on latents it was trained on (or generated itself). This is different from the formulation in section~\ref{sec:progmask} as we do not run the full sampling procedure until time $t$, the sampled image $\hat{z}_t$ is instead computed by a single DDIM sampling step from the previous noised image $z_t$:
\begin{equation}
\hat{z}_t = \sqrt{\alpha_{t}}f_\theta(z_{t+1}, t+1)  + \sqrt{1 - \alpha_t} \epsilon_\theta(z_{t+1}, t+1),
\end{equation}

where both $z_t$ and $z_{t+1}$ are the noisy versions of the original image via Eq.~\ref{forward_process}. For simplicity, we define the discrepancy as the pixel-wise L1 distance and compute the discrepancy statistics as:
\begin{equation}
\begin{split}
% \tilde{D_t} &= \left\lvert f_\theta\left(\hat{z}_t, t)\right) - f_\theta(z_t, t) \right\rvert,\\
\mu_{D_t} &= \mathbb{E}_{z\sim \mathcal{X}} \left[D_t(z_t,\hat{z}_t)\right], \\
\quad \sigma_{D_t}& = \mathbb{E}_{z\sim \mathcal{X}}\left[\sqrt{(D_t(z_t,\hat{z}_t) - \mu_{D_t})^2}\right].
% \mu_{D_t} &= \frac{1}{K} \sum_{i=1}^K D_t^{(i)}, \\
% \quad \sigma_{D_t}& = \sqrt{\frac{1}{K} \sum_{i=1}^K (D_t^{(i)} - \mu_{D_t})^2}.
\end{split}
\end{equation}
Note that $\mu_{D_t}$and $\sigma_{D_t}$ are both scalar values and represent the average expectation over all dimensions. The time-dependent threshold is subsequently defined as:
\begin{equation}\delta_t = \mu_{D_t} + \lambda\sigma_{D_t},\end{equation}

where $\lambda$ is a tunable hyperparameter that is constant across all time-steps.

\begin{figure}[tb]
  \centering
  \includegraphics[width=\linewidth]{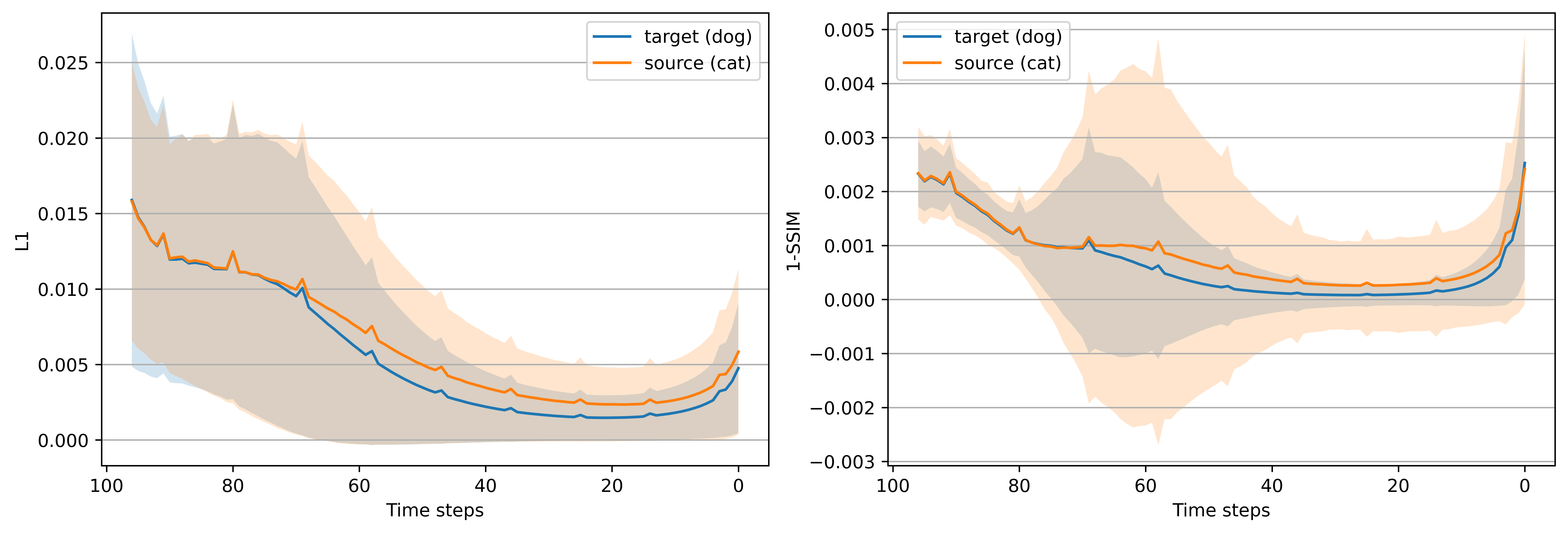}
  \caption{
  Mean and standard deviations for L1 (left) and 1-SSIM (right) discrepancy metrics. We compare statistics for reference from the target dataset (AFHQ-Dog) and a source dataset that is unknown to the diffusion model (AFHQ-Cat). Note that source dataset statistics are not used in our method, and are displayed only for presentation purposes.
  }
  \label{fig:D_t_stats}
\end{figure}
This allows for a dynamic thresholding scheme that is easily adjustable by a single parameter. In practice, we compute $\mu_{D_t}$ and $\sigma_{D_t}$ once for a given generative model and target distribution. Once computed, the distance statistics can be applied to mask generation for reference images from any source distribution. To ground our approach, in Figure~\ref{fig:D_t_stats},  we display plots of $\mu_{D_t}$ and $\sigma_{D_t}$ computed over the target distribution and a source distribution that the model was not trained on. It can be observed that predictions deviate the most around the middle of the sampling process. This is expected, as it is known that the middle time-steps represent the critical phase where global semantic structure and coarse geometry are resolved.

\subsection{Progressive Inpainting}

Although the masking framework described in Section~\ref{sec:progmask} employs the full DDIM generative process, it is not adequate to generate realistic samples from the target domain $\mathcal{X}$. After computing the sequence of masks $[M_T, M_{T-1}, ... M_1]$, we apply a modified version of the RePaint~\cite{lugmayr2022repaint} inpainting algorithm. In Repaint, time is periodically reversed so that denoising steps are repeated several times to allow the generative process to create a globally consistent image. However, instead of a constant mask, at each time-step $t$ of the backward diffusion process, we apply the corresponding mask $M_t$. As in previous works, we apply a slight dilation and Gaussian blurring filter to the mask, as it has been shown to improve image coherence at the mask boundaries~\cite{couairon2022diffedit}. Additionally, for better generation results, the inpainting can be run for a different number of total sampling steps $\tau$. In this case, we adapt the time schedule by utilizing the closest corresponding mask to the current time step $t$ as $M_{\lfloor t\frac{T}{\tau}\rfloor}$.

\section{Experiments}
In this section, we first introduce the Implementation details and setup for our experiments. Then, we present qualitative and quantitative comparisons with previous methods in unsupervised image translation. Lastly, we conduct ablation studies on the individual components and parameter values of our method.

\begingroup
\renewcommand{\arraystretch}{0.1}
\renewcommand{\topfraction}{1.}    % Allow up to 90% of top of page for floats
\renewcommand{\bottomfraction}{1.} % Allow up to 80% of bottom of page
\renewcommand{\textfraction}{0.05}  % Allow as little as 5% text on a page
\renewcommand{\floatpagefraction}{1.} % Requires 80% of page to be filled to put float on its own page

\begin{table}[t]
\centering
% \fontsize{8pt}{8pt}\selectfont
\caption{Quantitative comparison of our results with baseline methods on AFHQ and Celeba-HQ I2I translation tasks. ILVR, SDEdit, EGSDE, CycleDiffusion all use the same backbone generator model, the best result among these methods is denoted in bold. Method denoted with $\dagger$ require specialized training on the source domain. Our method outperformed baseline methods while remainging source agnostic.}\label{tab_i2i_comp}
% \rowcolors{4}{lightgray}{white} 
\begin{tabular}{lc@{\hspace{5pt}}c>{\hspace{5pt}}c>{\hspace{5pt}}c>{\hspace{5pt}}c}
\toprule

 \multicolumn{1}{c}{Method} && \multicolumn{1}{c}{FID $\downarrow$} {\hspace{5pt}}& \phantom{e-0}KID$\times10^3\downarrow$& SSIM $\uparrow$ & LPIPS $\downarrow$  \\
\midrule

\multicolumn{6}{c}{Wild $\rightarrow$ Dog}  \\
  
 \cmidrule(lr){1-6}
CUT$^{\dagger}$~\cite{park2020contrastive}&& 92.94&48.0&0.592&- \\
Santa$^{\dagger}$~\cite{xie2023unpaired}&&74.93&31.6&0.453&-\\
UNSB$^{\dagger}$~\cite{kim2023unpaired}&&86.72&40.2&0.524&-\\
SDDM$^{\dagger}$~\cite{sun2023sddm}&& 57.38  & - & 0.328 & - \\
\cmidrule(lr){1-6}
ILVR~\cite{choi2021ilvr}&& 81.34 ± 1.44 & 36.6 ± 1.56  & 0.289 ± 0.001 & 0.530 ± 0.001 \\
SDEdit~\cite{meng2021sdedit}&& 69.38 ± 1.10 & 27.6 ± 1.1  & 0.343 ± 0.001 & 0.515 ± 0.001 \\
EGSDE$^{\dagger}$~\cite{zhao2022egsde}&& 57.89 ± 0.62 & 19.2 ± 0.98 & 0.363 ± 0.001 & 0.504 ± 0.001 \\ 
CycleDiffusion$^{\dagger}$~\cite{wu2023latent}&& 56.10 ± 0.59 & 19.5 ± 1.02 &\setBold[0.4]0.479 ± 0.001\unsetBold  & 0.465 ± 0.001 \\
% \hline
 PM-Edit (Ours) && \setBold[0.4]55.64  ± 0.61\unsetBold & \setBold[0.4]17.2  ± 0.92\unsetBold &\setBold[0.4]0.479 ± 0.001\unsetBold   & \setBold[0.4]0.445 ± 0.001\unsetBold  {\hspace{5pt}}\\
 % progMask (Ours) && 57.66 $\pm$ 0.00 & 19.3 $\pm$ 0.00 &0.461 $\pm$ 0.00  &0.449 $\pm$ 0.00 {\hspace{5pt}}\\
 % progMask (Ours)&& 57.66 $\pm$ 0.00 {\hspace{5pt}}& 0.019 $\pm$ 0.00 {\hspace{5pt}}&0.46 $\pm$ 0.00 &&57.66 $\pm$ 0.00 {\hspace{5pt}}& 0.019 $\pm$ 0.00 &0.46 $\pm$ 0.00\\
 \midrule
\multicolumn{6}{c}{Cat $\rightarrow$ Dog}  \\
 \cmidrule(lr){1-6}
% & 78.52 & 33.3 & 0.359 &0.508
NOT$^{\dagger}$~\cite{korotin2022neural}&& 161.54&-&0.566&- \\
DIOTM$^{\dagger}$~\cite{choiimproving}&& 74.70&-&0.363&- \\
ASBM$^{\dagger}$~\cite{gushchin2024adversarial}&& 91.40&-&0.463&- \\
DSBM$^{\dagger}$~\cite{shi2023diffusion}&& 100.08&-&0.532&- \\
EGEOT$^{\dagger}$~\cite{mokrov2023energy}&& 53.29&-&0.349&- \\
\cmidrule(lr){1-6}
ILVR~\cite{choi2021ilvr}&& 75.38 ± 1.49 & 31.5 ± 1.95 & 0.359  ± 0.001 & 0.507 ± 0.001\\
SDEdit~\cite{meng2021sdedit}&& 73.72  ± 1.01 & 28.1 ± 1.13 & 0.418 ± 0.001 & 0.495 ± 0.001 \\
EGSDE$^{\dagger}$~\cite{zhao2022egsde}&& 63.37 ± 0.71 & 21.4 ± 1.21 & 0.437 ± 0.001& 0.471 ± 0.001 \\
CycleDiffusion$^{\dagger}$~\cite{wu2023latent}&& \setBold[0.4]58.87 ± 0.69\unsetBold & 20.4 ± 1.10&\setBold[0.4]0.557 ± 0.001\unsetBold& 0.426 ± 0.001 \\
% \hline
 PM-Edit (Ours)&& 61.31 ± 0.73 & \setBold[0.4] 18.3 ± 1.80 \unsetBold&0.551 ± 0.001  &\setBold[0.4]0.415 ± 0.001\unsetBold {\hspace{5pt}}\\
 % progMask (Ours)&& 55.40 $\pm$ 0.00 & 14.5 $\pm$ 0.00 &0.486 $\pm$ 0.00  &0.473 $\pm$ 0.00 {\hspace{5pt}}\\
 \midrule
 
\multicolumn{6}{c}{Male $\rightarrow$ Female}  \\
 
 \cmidrule(lr){1-6}
ILVR~\cite{choi2021ilvr}&& 60.17 ± 0.33 & 47.30 ± 1.22 & 0.510 ± 0.001 &0.390 ± 0.001 \\
SDEdit~\cite{meng2021sdedit}&& 54.91 ± 0.47 & 47.70 ± 1.34 & 0.577 ± 0.001 & 0.361 ± 0.001 \\
EGSDE$^{\dagger}$~\cite{zhao2022egsde}&& 48.50 ± 0.24 & 45.96 ± 0.96 & 0.579 ± 0.001 & 0.357 ± 0.001 \\
CycleDiffusion$^{\dagger}$~\cite{wu2023latent}&& \setBold[0.4]45.04 ± 0.23\unsetBold  & 45.11 ± 0.98 &0.564 ± 0.001& 0.369 ± 0.001\\
 PM-Edit (Ours)&& 50.99 ± 0.25 & \setBold[0.4] 42.10 ± 0.91\unsetBold & \setBold[0.4]0.586 ± 0.001 \unsetBold & \setBold[0.4]0.356 ± 0.001\unsetBold {\hspace{5pt}}\\
 % progMask (Ours)&& 57.66 $\pm$ 0.00 {\hspace{5pt}}& 0.019 $\pm$ 0.00 {\hspace{5pt}}&0.46 $\pm$ 0.00 &&57.66 $\pm$ 0.00 {\hspace{5pt}}& 0.019 $\pm$ 0.00 &0.46 $\pm$ 0.00\\
 \bottomrule
 
\end{tabular}
\end{table}

\begin{table}[tb]
      % \fontsize{8pt}{8pt}\selectfont
      
        \caption{Results on the PIE-Bench (Prompt-driven Image Editing Benchmark). Our method outperform previous methods in both fidelity in prompt adherence.}
        % \vspace{-17pt}
      \label{tab:PIEbench}
      \centering
      \scalebox{1}{
      \begin{tabular}{@{}l|c|cc|ccc@{}}
        \toprule
            &  \textbf{Structure}&  \multicolumn{2}{c|}{\textbf{Fidelity}}& \multicolumn{2}{c}{\textbf{CLIP Similarity}}\\
        \midrule
          \textbf{Method}& \textbf{Distance$_{\times10^3}\downarrow$ }& \textbf{LPIPS$_{\times10^3}\downarrow$}&
         \textbf{SSIM$\uparrow$}&
         \textbf{Whole$\uparrow$}& \textbf{Edited$\uparrow$}\\
        \midrule
         MasaCtrl.~\cite{cao2023masactrl} &  24.70 & 87.94 &  0.813 & 24.38 & 21.35 \\
         % DiffEdit &  5.04 & 29.30 &  0.875 & 23.61 & 20.34\\
         Wang et al.~\cite{wang2024unified} &  22.40 & 84.45 &  0.816 & 25.15 & 22.12 \\
         LatentEdit~\cite{liu2025latentedit} & 22.13& -& 0.808& 25.45& 22.51 \\
         PnP~\cite{tumanyan2023plug} & 28.22 & 113.46 & 0.791 & 25.41 & 22.55\\
         PnPInversion~\cite{ju2024pnp} & 24.29 & 106.06 & 0.797 & 25.41 & 22.62\\
         PM-Edit (Ours) & \setBold[0.4]19.68\unsetBold & \setBold[0.4]83.51\unsetBold & \setBold[0.4]0.819\unsetBold & \setBold[0.4]26.43\unsetBold & \setBold[0.4]22.81\unsetBold \\
      \bottomrule
      \end{tabular}  
      }
\end{table}

\endgroup

\begin{figure}[tb]
  \centering
  \includegraphics[width=1.\linewidth]{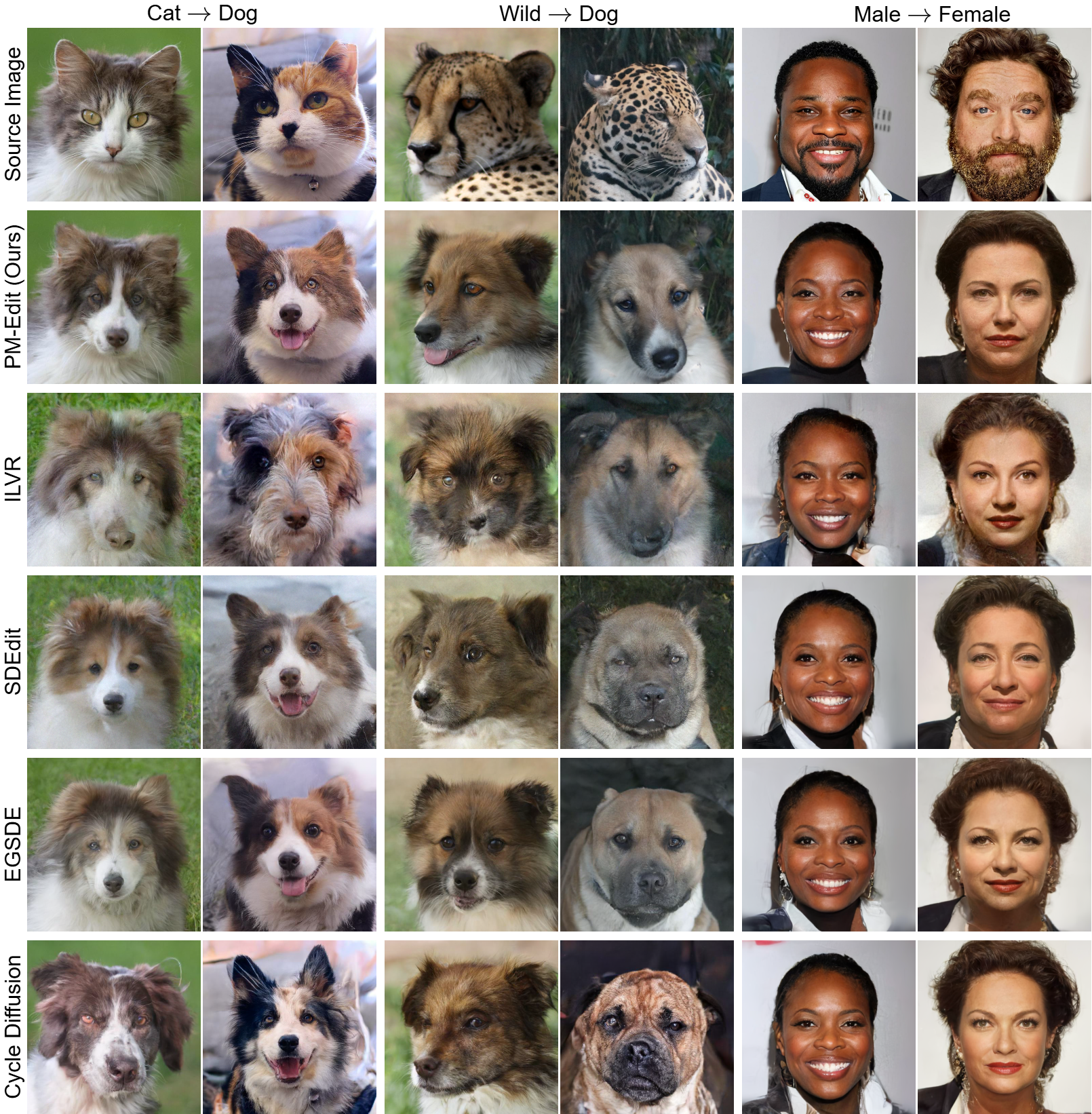}
  \caption{
  Qualitative comparison of our method and other diffusion-based  approaches.
  }
  \label{fig:i2i_comp}
\end{figure}

\subsection{Implementation Details}
\subsubsection{Datasets.} We conduct I2I translation experiments using two common benchmark datasets: AFHQ~\cite{choi2020stargan} and Celeba-HQ~\cite{karras2017progressive}. AFHQ contains high quality images divided into three categories: cat, dog, and wild; we evaluate on the Cat$\rightarrow$Dog and Wild$\rightarrow$Dog translation tasks. We only use the test split of the dataset; it contains 500 test images for each category. Celeba-HQ consists of high-resolution images split into two categories: male and female. We evaluate on the Male$\rightarrow$Female task, using the validation split of 1000 images. For both datasets, images are resized to 256$\times$256. We also evaluate our method on the PIE-Bench~\cite{ju2024pnp} dataset for text prompt guided multi-aspect image editing. It consists of 700 images, each assigned a source and editing prompt, encompassing a diverse range of manipulations, from object edits to style transfer tasks.

\begin{figure}[tb]
  \centering
  \includegraphics[width=\linewidth]{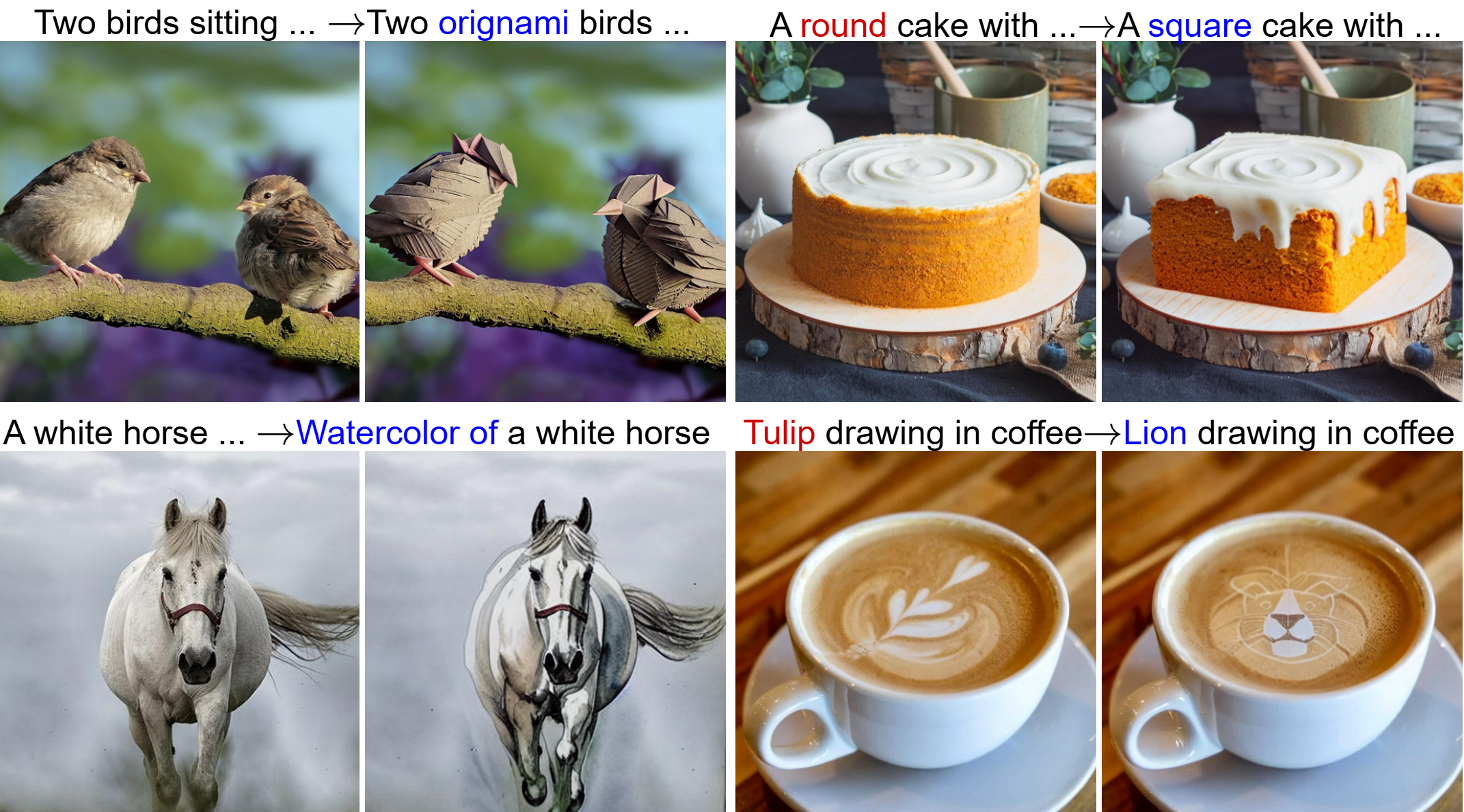}
  \caption{
  Examples of text prompt guided Image translation on the PIE-Bench dataset. The left image and text for each sample represents the source image-prompt pair. With the result given by our method, according to the target prompt, shown on the right.
  }
  \label{fig:pie_bench}
\end{figure}

\begin{figure}[tb]
  \centering
  \includegraphics[width=\linewidth]{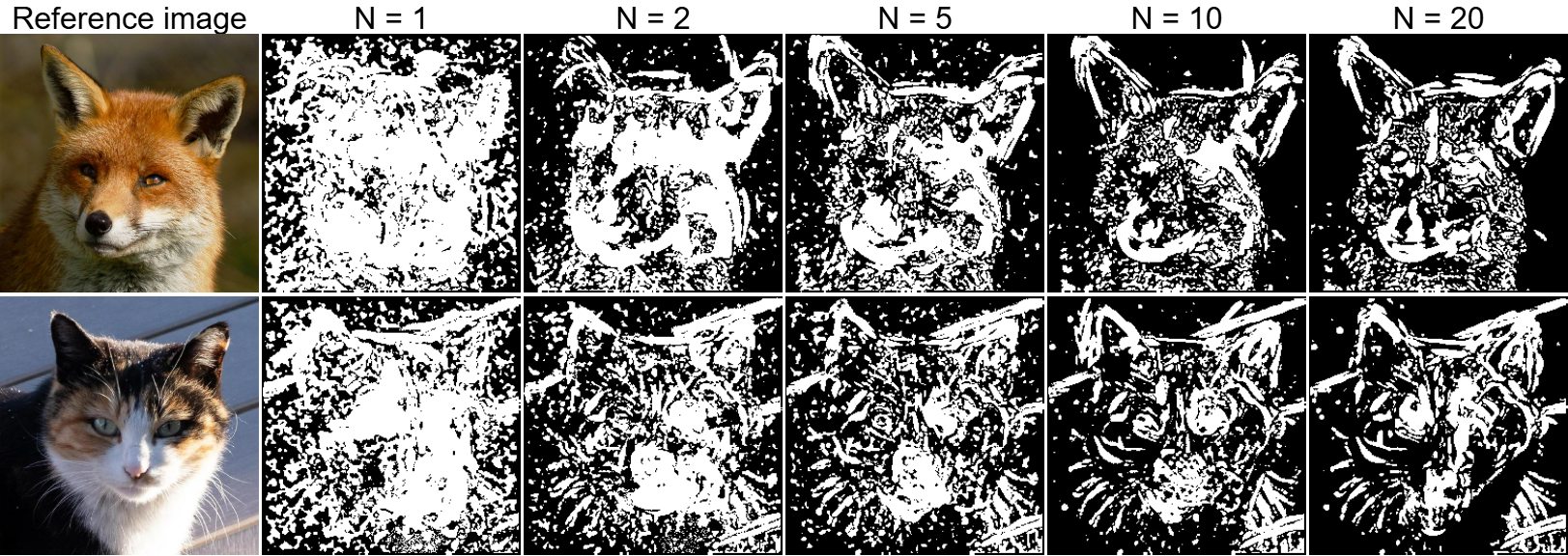}
  \caption{
  Demonstration of the final mask for different numbers of mask-computation batches $N$. Computing discrepancies over of average of several images yields cleaner masks with less noise interference.
  }
  \label{fig:batch_size}
\end{figure}

\subsubsection{Evaluation Metrics.} To properly evaluate image translation, we measure two aspects: realism and faithfulness. For realism, we compute Frechet Inception Distance (FID)~\cite{heusel2017gans} and Kernel Inception Distance (KID)~\cite{binkowski2018demystifying} scores. Following CUT~\cite{park2020contrastive}, on AFHQ they are computed between the generated images and the target test split, while for Celeba-HQ we use the target train split like in EGSDE~\cite{zhao2022egsde}.  To evaluate faithfulness, we report Structural Similarity Index Measure (SSIM)~\cite{wang2004image} and Learned Perceptual Image Patch Similarity (LPIPS)~\cite{zhang2018unreasonable} between the generated samples and their respective reference images. In this setting, LPIPS and KID are considered more reliable metrics, as FID is known to be inaccurate for small datasets, and SSIM does not capture human level perception as well as LPIPS. For PIE-Bench we evaluate prompt-image consistency using CLIPSIM~\cite{wu2021godiva} and fidelity via structure distance~\cite{tumanyan2022splicing}.

\subsubsection{Implementation Details.} On the unconditional I2I translation tasks, we utilize the same pretrained diffusion backbones as in previous works. We use the AFHQ dog generator network provided by ILVR ~\cite{choi2021ilvr} and the Celeba-HQ network provided by the authors of EGSDE. Both networks were trained only on the target domain, and we do not fine tune either model. When computing the masks, we set the threshold coefficient to be $\lambda=1.2$ and use $N=10$ image batches. During inpainting, we first dilate and then blur the masks with a kernel size of 11 in both, and set the gaussian blur sigma to be 5.0. We generate the mask over 100 steps of a DDIM backward process but perform inpainting over the full 1000 steps. We integrate Repaint~\cite{lugmayr2022repaint} by reverting back for 10 time steps at every 20th step, and repeat this action twice. We use $1-SSIM$ for the discrepancy metric $D_t$, and have found $\mu_{D_t}$ and $\sigma_{D_t}$ to be robust across models and datasets. We use $\mu_{D_t}$ and $\sigma_{D_t}$ computed on a model trained on AFHQ-dog in all presented results. Other configurations are explored in the Supp. Material.

On PIE-Bench, we used Sable Diffusion v1.5~\cite{podell2024sdxl} as a backbone and applied our method in the latent space of the VAE with L1 as the discrepancy metric. Performing inpainting in an LDM's latent space does not require extra RePaint steps thanks to the robustness of the VAE decoder; this significantly lowers execution time, which we explore in Sec. D.2 of the Supplementary Material. Furthermore, in our PIE-Bench experiment, we first perform DDIM inversion using the source prompt of a given sample; this improves faithfulness and is standard in prompt guided image translation methods.

\subsection{Image to Image Translation}
 We provide a quantitative comparison of our method with previous baseline works in Table~\ref{tab_i2i_comp}. Our progressive masking and editing via inpainting pipeline is denoted as PM-Edit. It outperforms the other source-free approaches in ILVR~\cite{choi2021ilvr} and SDEdit~\cite{meng2021sdedit}; and achieves competitive results with methods that rely on training models using the source domain. Notably, we observe lower scores in both KID and LPIPS, which are considered to be the more accurate metrics. KID is a better metric than FID when used on small datasets; and LPIPS is considered to have better correlation with human perception than SSIM. Qualitative comparisons are available in Figure~\ref{fig:i2i_comp}, showing that our masking approach has an advantage in preserving class-irrelevant details without hindering realism. 
 
 Quantitative comparisons for PIE-Bench are shown in Table~\ref{tab:PIEbench}. Our method represents a solid balance between prompt adherence and faithfulness to the source image, reflected by superior results across all metric compared to other recent methods. Qualitative results in Figure~\ref{fig:pie_bench} show examples of text guided image editing. Thanks to the masking-based approach our method performs minimal changes to the source image by preserving image areas that are not relevant to the editing prompt.

% \subsection{Anomaly Detection}

\subsection{Ablation Study}
In Table~\ref{tab:lambda}, we present results for different values of the threshold coefficient $\lambda$. A higher threshold value leads to stricter masking, which yields higher SSIM and LPIPS scores. Conversely, lower $\lambda$ values lead to better FID and KID scores because the generation process is less restricted. This means that by adjusting $\lambda$ we gain a tradeoff between realism and faithfulness in the translation. Figure~\ref{fig:batch_size} shows the effect of varying values of the mask computation batch size $N$. Computing the prediction discrepancy over only one random noise seed is sensitive to outliers, and yield to a noisy mask. Using a larger number of images in a batch increases our method's robustness to these outliers, but comes at the cost of higher computation time. We include a detailed analysts of run-time efficiency in section D.2 of the Supplementary Material.

In Table~\ref{tab:settings} we present ablations for the different components of our method. First we we compute the mask normally, but do not use dynamic masking during inpainting. In the \emph{"constant $M_{0.5T}$"} the mask computed at the middle, and in \emph{"constant $M_0$"} only the final mask is used. These settings are similar to the constant masking in DiffEdit~\cite{couairon2022diffedit}, but neither show comparable results to our dynamic mask inpainting approach. This is because a larger mask at smaller $t$ unnecessarily covers region of an image at early steps, this means that some structural features might not be preserved, or insignificant details are lost. In contrast, computing the mask too early limits the generation too much, which leads to unrealistic final image that does not resemble the target domain.
 
Table~\ref{tab:settings} also contains ablations that justify our adaptive masking and inpainting approach by experimenting with a constant threshold set to the equivalent of $\lambda=1.2$ at $t=0.5T$ across the entire sampling process, labeled as \emph{"constant $\delta_t$"}. This fails to generate translations faithful to the original images, as discrepancies at the early steps make the mask too large. Additionally, we ablate for when a single batch is used during mask computation ($N=1$). Which shows that as statistical errors dominate the discrepancy computation, the mask ends up covering too much of the image and lowers faithfulness to the reference image. Finally demonstrate results of our method without utilizing the Repaint inpainting technique. Although the LPIPS score in this setting is similar to our default setting, the KID of the resulting images is low. This shows that applying RePaint is highly beneficial for realism while not significantly hindering faithfulness.

\begin{table}[tb]
    \centering
    % \parbox{0.49\textwidth}{
    %     \caption{Running time (seconds) and DICE score comparison of the different methods and solver steps. 
    %     }
    %     % \rowcolors{2}{lightgray}{white} 
    %     \begin{tabular}{l>{\hspace{10pt}}c>{\hspace{5pt}}c>{\hspace{5pt}}c}
    %         \toprule 
    %         Method & I-AUROC $\uparrow$ & P-AUROC $\uparrow$\\
    %         \midrule
    %         1 & 0.8034 & 30.95 \\ %30.95
    %         3 & 0.8036  & 39.86\\ %40.74
    %         5 & 0.8038  & 58.17\\%60.08
    %         10 & 0.8039  & 98.68\\
    %         20 & 0.8105  &  251.64\\
    %         \bottomrule
    %         \end{tabular}
    %     \label{tab:times}
    % }
    \parbox{0.49\textwidth}{
        \caption{Ablation study for different modifications to our baseline approach on Wild$\rightarrow$Dog. In each test other components are set to default unless ablated for.
        }
        % \rowcolors{2}{lightgray}{white} 
        \begin{tabular}{l>{\hspace{10pt}}c>{\hspace{5pt}}c}
            \toprule 
            Setting & KID $\downarrow$ &  LPIPS $\downarrow$\\
            \midrule
            constant mask $M_{0.5T}$ & 33.3 & 0.309\\
            constant mask $M_{0}$ & 13.1 & 0.533\\
            constant $\delta_t$ & 10.5 & 0.698\\
            w/o batching (N=1) & 16.2 & 0.617\\
            % w/o blurred mask & 0.8034 & 30.95\\
            w/o RePaint & 28.0 & 0.4373 \\ 
            \bottomrule
            \end{tabular}
        \label{tab:settings}
    }
    \hfill
    \parbox{0.48\textwidth}{
        \caption{Ablation study for different values of the threshold coefficient hyperparameter $\lambda$ on Wild$\rightarrow$Dog. We observe a tradeoff between realism and faithfulness. 
        }
        % \rowcolors{2}{lightgray}{white} 
        \begin{tabular}{l>{\hspace{10pt}}c>{\hspace{5pt}}c>{\hspace{5pt}}c>{\hspace{5pt}}c>{\hspace{5pt}}c}
            \toprule 
            $\lambda$ & FID $\downarrow$ & KID $\downarrow$& SSIM$\uparrow$ & LPIPS$\downarrow$\\
            \midrule
            1.0 & 54.12 & 17.1 & 0.453 &  0.459\\ %30.95
            1.2 & 55.64& 17.2 &0.479&0.445 \\ %40.74
            1.4 & 58.34& 19.4&0.497& 0.442\\%60.08
            1.6 & 61.35  &  21.4 & 0.521 & 0.413\\
            1.8 & 62.89  &  22.6 & 0.531& 0.406\\
            \bottomrule
            \end{tabular}
        \label{tab:lambda}
    }
\end{table}

\section{Conclusion}

In this paper, we present a novel source-agnostic framework for unsupervised image-to-image translation. By leveraging the internal dynamics of denoising diffusion models, our method overcomes the limitations of static masking and the requirement for source-domain supervision. We achieve robust dynamic masking by introducing a time-dependent, statistical thresholding mechanism. By dynamically isolating domain-specific discrepancies while preserving invariant structural features, our method ensures faithfulness to the original content without compromising the realism of the target transformation. Extensive experiments on the AFHQ and Celeba-HQ datasets demonstrate that our method consistently outperforms existing state-of-the-art unsupervised methods in both realism and faithfulness while not integrating knowledge about the source domain. Because our method requires only a pretrained model of the target domain and no additional training or source-domain data, it offers a highly flexible solution for image translation. However, a current limitation of our approach is the underlying assumption that the source and target domains are of the same modality. Future work could explore implementing our method using more sophisticated discrepancy metrics that could be adaptable to cross-domain translation.

\section*{Acknowledgement}
This work was supported by the Korea government (MSIT): IITP-RS-2021-II211341, Artificial Intelligence Graduate School, Chung-Ang University and NRF-RS-2025-25462275.
\bibliographystyle{splncs04}
\bibliography{main}
\end{document}